# Hallucination Level of Artificial Intelligence Whisperer – Case Speech Recognizing Pantterinousut Rap Song

Ismo HORPPU[1], Frederick AYALA[2], Erlin GULBENKOGLU[3], Mc Timo, and Syntikka Janne

Email: ishorppu@gmail.com[1], frederickayala@gmail.com[2], erlingulbenk@gmail.com[3]

**Abstract**: All languages are peculiar. Some of them are considered more challenging to understand than others. The Finnish language is known to be a complex language. Also, when languages are used by artists, the pronunciation and meaning might be more tricky to understand. Therefore, we are putting AI to a fun, yet challenging trial: translating a Finnish rap song to text. We will compare the Faster Whisperer algorithm and YouTube's internal speech-to-text functionality. The reference "truth" will be Finnish rap lyrics, which the main author's little brother, Mc Timo, has written. Transcribing the lyrics will be challenging because the artist raps over synth music played by Syntikka Janne. The hallucination level and mishearing of AI speech-to-text extractions will be measured by comparing errors made against the original Finnish lyrics. The error function is informal but still works for our case.

**Keywords:** AI, ASR, Hallucination, Faster Whisperer, speech-to-text, Finnish rap, Pantterinousut.

**Disclaimer:**
*This paper has nothing to do with our current or past employers. In addition, this work is done a bit like tongue-in-cheek although the first two authors have Ph.D degrees.*

Artificial Intelligence has been used while writing this paper. Specifically, Google Gemini 2.5 Pro to summarize available speech-to-text services, ChatGPT to help in defining error function (for real lyric versus ASR heard one), and LALAL.AI has been applied as one of the data processing algorithms. However, the authors think that most of this paper is human-powered.

## 1. Introduction and Problem Description

There has been lots of hype about artificial intelligence and concerns about hallucinations (and biases). Ideally, we want to get subtitles created automatically for a song with Finnish rap lyrics and instrumental background music. Finnish was picked as it is not available in all speech-to-text services and models, and it is more difficult than some other languages.

*1.1 – Problem Description*

We want to translate Finnish rap lyrics automatically to text format while minimizing the hallucination of AI (= crazy text extractions from rap lyrics). The input data source will be an audio track from the Pantterinousut video published by my little brother Mc Timo and Syntikka Janne (= keyboard player) in June 2025. This song is available on YouTube (see ref. 1.). In addition, there is also a Facebook page for the band (ref. 2.). Figure 1 depicts the start of the video, where the Finnish word "pelaa" means play.

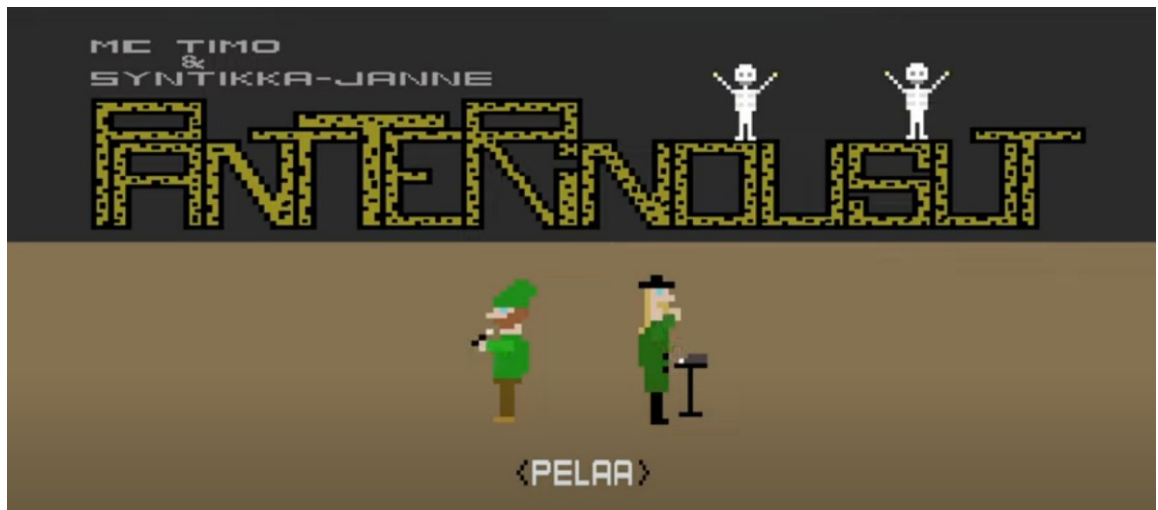

Figure 1. Start of the Pantterinousut video (from which we want to extract the Finnish rap lyrics automatically).

## 3. Speech-to-text Models and Model Selection

There are both online (cloud services mostly) and offline (models run locally) approaches for translating speech-to-text. Below we will summarize services.

*3.1 – Available online and offline models*

There are many online and offline services available for such already. Namely, online options include services such as: VEED.IO, Submagic, Descript, Kapwing, and Google's Cloud speech to text service (or Microsoft's or Amazon's similar ones).

However, we don't want to do subtitles online. Instead, we prefer doing them locally using an offline AI model. At least three software can create subtitles automatically and locally. These include LLPlayer (ref. 7.), POTPlayer (ref. 8.), and VLC MediaPlayer (ref. 10.). The last of them introduced free AI subtitles (and real-time translation) with support for 100+ languages, ref. 4. However, the authors of this article could not find even an experimental build of VLC Media Player that would enable this feature (experimental builds are available from ref. 5.).

*3.2 – Models that we are going to use and evaluate*

We will use the Faster Whisperer model (ref. 3.) and YouTube's integrated speech-to-text algorithm. However, we do not know details (= what it has eaten) for the latter model. Faster Whisper is a reimplementation of OpenAI's Whisper (ref. 6.) model using CTranslate2. That is a fast inference engine for transformer models.

*3.3 – Hypothesis: which model is going to be the winner?*

Our hypothesis or "prior information" is that we expect YouTube's automatic text-to-speech translation to be better than the Faster Whisperer algorithm. We have a gut feeling about that, and it is also a realistic expectation as YouTube likely has allocated quite some R&D resources for that. It is essential for YouTube. Namely, YouTube video completion rates grow significantly with subtitles, ref. 23.

*3.4 – Model Selection*

As data scientists, we typically favor employing methods such as the Akaike Information Criterion (AIC) model selection criterion. However, applying this criterion to our data presents a challenge. Fortunately, there are certain rules of thumb available. Notably, Sanseviero (ref. 9) shared a selection logic in late 2023 (which remains quite effective). The model selection logic is presented in Figure 2. In our specific scenario, the audio is not English but Finnish. Consequently, the decision tree for model selection suggests selecting the Whisper V3 large model.

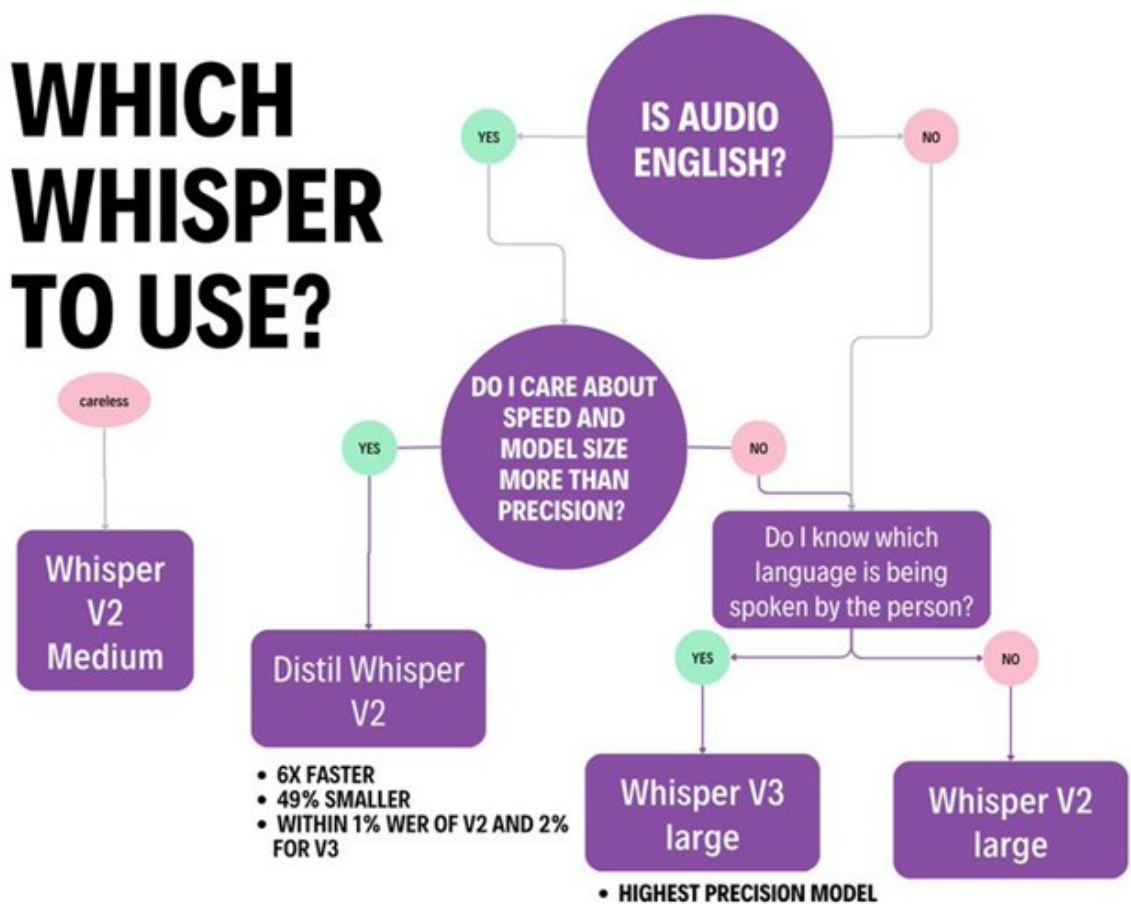

Figure 2. Rules for selecting a suitable Whisper model, image by Omar Sanseviero (ref. 9.)

## 4. Architecture of OpenAI Whisper Model

Figure 3 shows the architecture of OpenAI's Whisper model. It is a beautiful model, like most neural network models. More formally, it is an encoder-decoder transformer. The encoder takes in audio in 30-second chunks and creates a logarithmic spectrogram. Non-scaled spectrograms place more emphasis on higher frequencies, whereas log transferring focuses on lower frequencies.

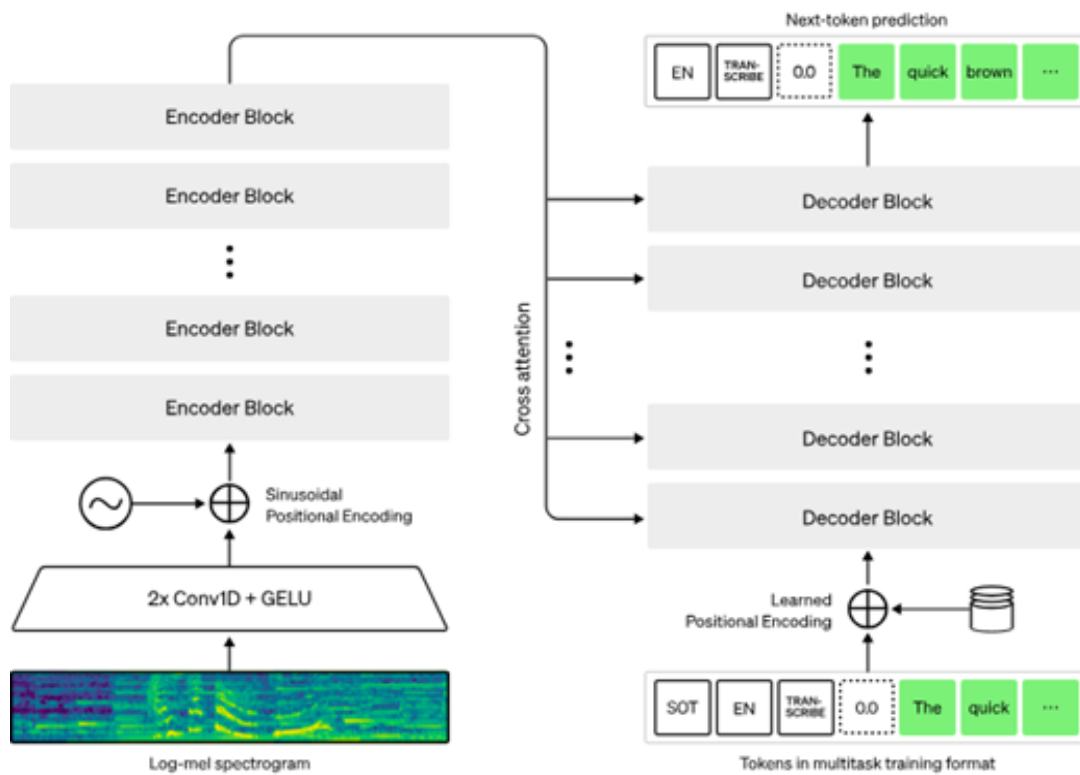

Figure 3. Architecture of OpenAI's Whisper model, image by OpenAI (ref. 6.)

As a reference, Figure 4 shows the spectrogram of the audio file from the Pantterinousut audio (for 2 minutes and 30 seconds). It shows how the frequency content of our audio signal changes over time. The spectrogram of audio was created using the below Python code. Figure 5 presents the linear and logarithmic frequency analyses of the audio, done using Audacity.

```python
import os
import wave
import pylab

def draw_spectrogram(file):
    sound_info, frame_rate = get_wav_info(file)
    pylab.figure(num=None, figsize=(19, 12))
    pylab.subplot(111)
    pylab.title('Spectrogram of %r' % file)
    pylab.specgram(sound_info, Fs=frame_rate, cmap='plasma')
    pylab.savefig('pantterinousut_spectrogram.png')

def get_wav_info(file):
    wave_file = wave.open(file, 'r')
    audio_frames = wave_file.readframes(-1)
    sound_info = pylab.fromstring(audio_frames, 'int16')
    frame_rate = wave_file.getframerate()
    wave_file.close()
    return sound_info, frame_rate

draw_spectrogram('pantterinousut.wav')
```

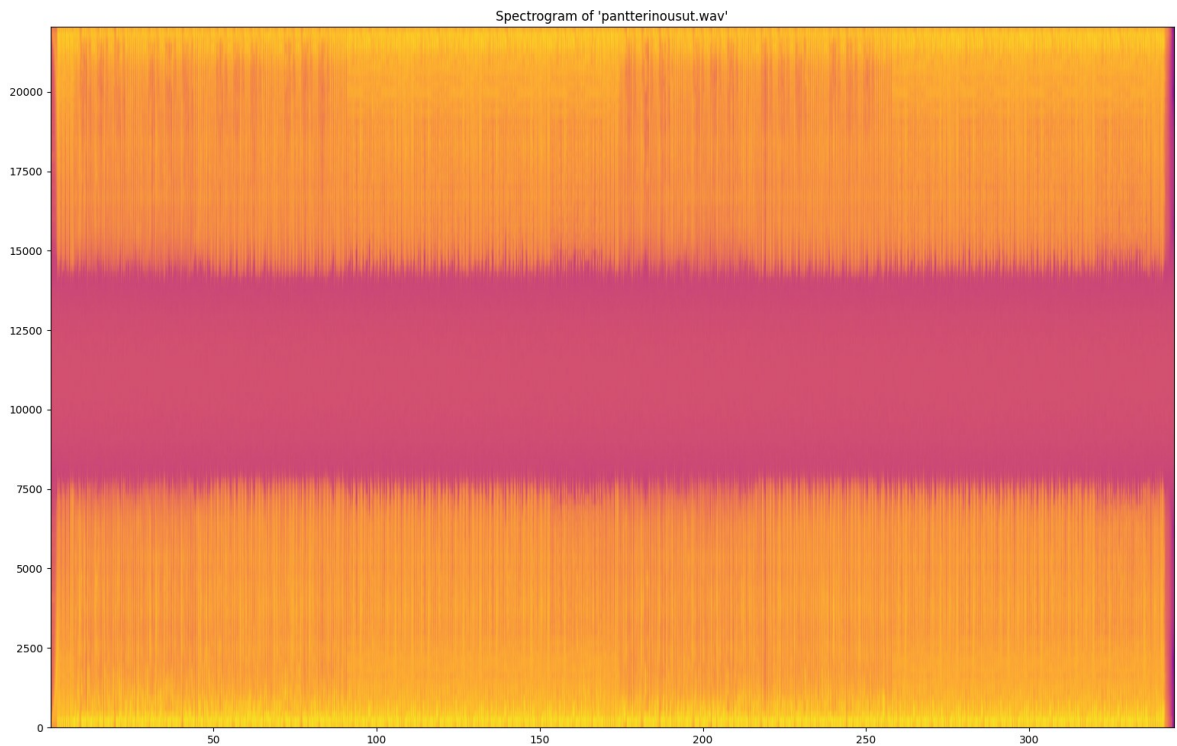

Figure 4. Spectrogram of audio of the Pantterinousut song (X is time and Y is frequency).

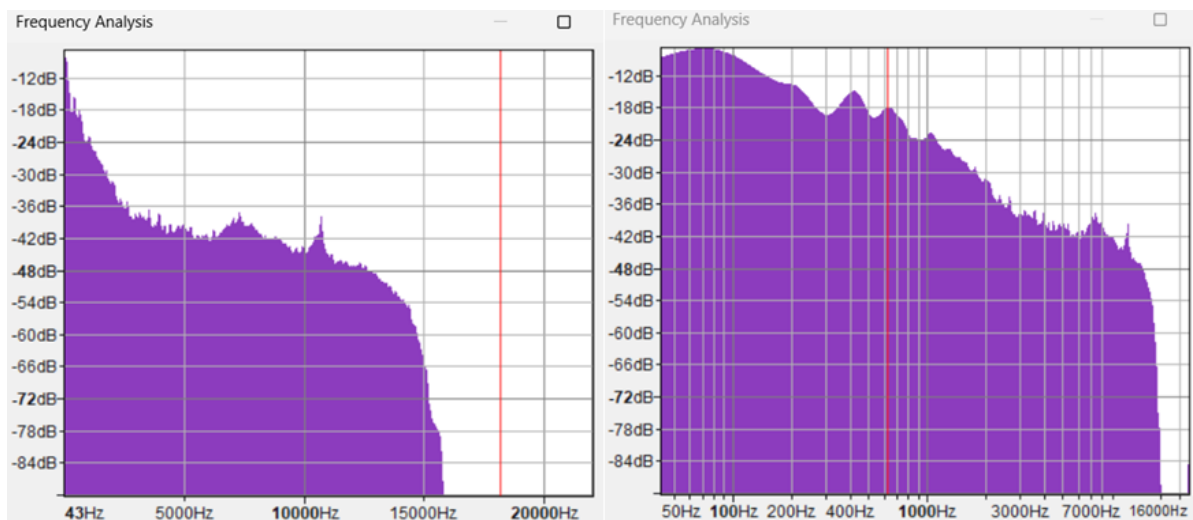

Figure 5. Linear and logarithmic frequency analyses done using Audacity, ref 18.

## 5. Speech-to-text Extraction Examples / The Faster Whisperer

We first tried to use ASR (= automatic subtitle creation) from LLPlayer and PotPlayer. However, we could not get them to work in the Windows 11 operating system. As a consequence, we just installed Nvidia CUDA (Intel I7 13700k with Nvidia 4070 RTX display card), then downloaded and executed the Faster Whisperer standalone version from the command prompt as:

```
Faster-whisper-xxl pantterinousut_audio.wav –language Finnish –model large –output_dir sourcewav
```

where pantterinousut_audio.wav is our audio file, the second parameter instructs the Whisperer model that the language is Finnish, and the last parameter tells which folder to write the output. It will create a .SRT subtitle file.

Execution of the Faster Whisperer took 1 minute and 33 seconds. So, extraction was faster than in real-time: the song and video lasted 2 minutes and 52 seconds. Figure 6 shows how the Faster Whisper model extracts rap lyrics from the song. We have highlighted one extraction with a white background. It says that between timestamps 00:34.700 and 00:39.940 Mc Timo rapped "Lakki lukkee lähikirjastosta" (roughly translates into English as "Lucky Luke from a local library"). An example of AI hallucination is the "Tordaita alas, niinku katu hauskaa" extraction. It should be "Portaita alas, niin kuin katuhaukka" which translates into English as "Down the stairs like Street Hawk".

Figure 6. Applying Faster Whisper XXL model to the Pantterinousut song.

## 6. AI Trial: Evaluation of Hallucination Levels

Next, we compare the speech-to-text algorithm of YouTube (we do not know how it has been implemented) versus the Faster Whisperer ASR model. We compare the performance for selected parts of the lyrics in the Pantterinousut song. The first author of the paper, a native Finnish speaker, is going to be the judge here. Hallucinations (speech-to-text errors) are marked below with a red background. We use the Finnish language because the original vocals contain rap rhymes in Finnish, and automatically translating them into English is difficult (we tried that). We include YouTube's automatic translation into English for the first case.

*6.1 – Used Error Metric*

We calculate the error for transcribed speech using the following implicit (human-assisted) approach. We check the following steps for each difference between translated speech and real lyrics:
1. Assess if the difference is hallucination or mishearing/incorrect prediction.
2. If it is mishearing, then calculate the Levenshtein string edit distance metric between misheard word(s) and real. E.g., the real Finnish lyric is "niin kuin" and the heard one is "niinu".
3. If mishearing is a single character with a low phonemic distance, we halve string edit distance. As an example, the real lyric is "Boston" and the heard one is "Poston": b<->p has a low phonemic distance. We expect that there can be ASR model hearing issues, at least with b<->p and t<->d phonemes, like for humans (see ref. 21).
4. The winner evaluation is the following:
   - If there is a hallucination, then the other model without a hallucination wins.
   - If both models hallucinate, then the one with less hallucination wins.
   - If both models have the same hallucination level, then one with a smaller string edit distance wins (= fewer misheard characters).
   - Else it is a draw.

One can ask which matters more: hallucination or mishearing. We decide that we penalize hallucination the most. However, in practice, a misheard word with considerable Levenshtein distance metric can also be bad when considering understanding heard rap.

*6.2 – Evaluation*

Below, we mark hallucinations with a red background color and misheard words with yellow. We evaluate our error metric per each case, and the value of the Levenshtein metric is included.

*Lyric #1*
Real:             **"…Kuka tykkäs JRstä oli täys pelle"**
YouTube:          "Kuka tykkäs JRstä oli täys pelle." (who liked JR was a complete clown)
Faster Whisperer: "Kuka tykkäs JRstä oli täys pelle."
Verdict:          Draw - both are correct (perfect extract).

*Lyric #2*
Real:             **"…jääkaapille kiireesti heti TV-pöllöstä"**
YouTube:          "jääkaapille kiireesti heti TV-pöllöstä"
Faster Whisperer: "jääkaapille kiireesti heti TV-pöllöstä"
Verdict:          Draw - both are correct (perfect extract).

*Lyric #3*
Real:             **"kaikki vihas spektran sallii"**
YouTube:          "kaikki vihas spektran salli"
Faster Whisperer: "kaikki viha spektran salliin"
Verdict:          YouTube misheard one word (Levenshstein distance 1), whereas Whisperer misheard two words (Levenshstein distance 1+2=3)
                  → YouTube wins.

*Lyric #4*
Real: **"Torspolla lämärin maaliin lataa"**
YouTube: "Torskolla lämärin maaliin lataa"
Faster Whisperer: "Torspolla lämärin maaliin lataa"
Verdict: YouTube misheard one character (Levenshstein distance 1) and Whisperer is perfect → Whisperer wins.

*Lyric #5*
Real: **"Portaita alas, niin kuin katuhaukka"**
YouTube: "Portaita alas niinu katu haukka"
Faster Whisperer: "Tordaita alas, niinku katu hauskaa"
Verdict: Faster Whisperer hallucinates twice and mishears one word (Levenshstein distance 3), Youtube mishears one word (Levenshstein distance 4) → Youtube is the winner.

*Lyric #6*
Real: **"Kovaa pelii, Bostoni palaa"**
YouTube: "Kovaa peli, Bostoni palaa"
Faster Whisperer: "Kovaa pelii, postoni palaa"
Verdict: Both models mishear one word (Levenshtein distance 1, halved for Whisperer) → Whisperer wins because hearing p instead of b (low phonemic distance).

*Lyric #7*
Real: **"Bubble bobble, hubba puppa"**
YouTube: "kuple pople kuppa puppa"
Faster Whisperer: "Puple pople, kuppa puppa"
Verdict: YouTube hallucinates twice (+ Levenshstein distance 3), whereas Whisperer hallucinated only once (+ Levenshstein distance 6) → Whisperer wins.

YouTube won twice out of seven times, whereas Faster Whisperer won thrice out of seven times (the draw happened twice). However, based on this simple and limited analysis, there is no statistical confidence in assessing which one of them is better.

## 7. Human helping AI: Pre-processing of Data

AI might have less difficulty interpreting audio vocals to text if we can pre-process the audio file. Pre-processing means filtering instrumental music away and leaving vocals only. Formally, we could try to decompose audio data at time point *t* using additive model formulation as: **audio(t) = vocals(t) + instrumental(t)**

We tried the following three approaches in our experiments to split audio using the above additive model formulation:
1) Statistician/IT engineer: "Let's do blind source separation (BSS). To be specific, apply Independent Component Analysis (ICA) with two latent components",
2) Sound engineer: "Let's cancel center out and hope for the best", and
3) Machine learning engineer: "Let's apply state of the art stem splitter".

**Approach #1: To be independent or not to be**

In ICA, we have multiple signals and observe mixtures of multiple original signals. See Figure 7 for the simplest case of a so-called cocktail party problem. Two original signals S1 and S2 are recorded (via linear mixing) to microphones X1 and X2. In our case, S1 is vocals and S2 is instrumental music. In addition, "microphones" are our audio track's left and right channels. Remark that the order and amplitude of signals are likely not preserved. In addition, only one of the original signals can be Gaussian (or decomposition fails). However, our audio track has no Gaussian signals (vocal track and instrumental track).

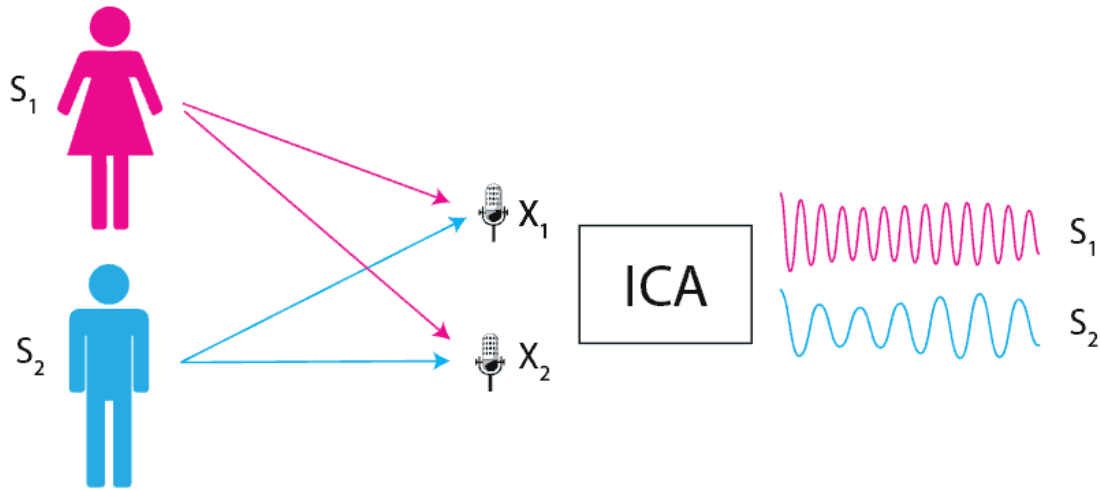

Figure 7. Idea of ICA in so-called cocktail party problem with two original signals, image by Enes Zvornicanin, ref 25.

We have a stereo audio file split into left and right channels. Then, we assume that these two channels separated measure two independent signals with a linear mixing. Ideally, these two signals are vocal and instrumental music. This sounds great on paper. As a statistician, the first author of this paper cannot avoid the temptation of getting independent data (as statisticians 'love' independent data points). We applied the FastICA algorithm invented by Hyvärinen (2009) and described by Hyvärinen and Oja (2010). It is worth mentioning that ICA itself has been studied by many Finnish researchers such as Taskinen, Sirkiä, Oja (2007, ref. 13.), Koivunen et al. (2008, ref. 15), and Ollila (2010, ref. 14).

Output of the algorithm is two beautiful independent signals. However, vocals and instrumentals are still mixed in both signals.

```python
# [INDEPENDENT COMPONENT ANALYSIS: APPROACH 1.]
import numpy as np
import wave
from sklearn.decomposition import FastICA
from scipy.io import wavfile

# Load in mono left and right channel WAV files.
mix_1_wave = wave.open('left.wav','rb')
framerate = mix_1_wave.getframerate()
signal_1_raw = mix_1_wave.readframes(mix_1_wave.getnframes())
signal_1 = np.frombuffer(signal_1_raw, np.int16)

mix_2_wave = wave.open('right.wav','rb')
signal_2_raw = mix_2_wave.readframes(mix_2_wave.getnframes())
signal_2 = np.frombuffer(signal_2_raw, np.int16)

# Initialize FastICA with two latent components.
ica = FastICA(n_components=2)
```

```python
# Run the FastICA algorithm using fit_transform on input data
ica_output = ica.fit_transform(np.c_[signal_1, signal_2])

# Convert to signed integer, map range, and amplify by 100x
result_signal_1_int = np.int16(ica_output[:,0] * 32767 * 100)
result_signal_2_int = np.int16(ica_output[:,1] * 32767 * 100)

# Write wave files
wavfile.write("signal_1.wav", framerate, result_signal_1_int)
wavfile.write("signal_2.wav", framerate, result_signal_2_int)
```

**Approach #2: Cancelling center**

Figure 8 depicts a bit of an idea to cancel the center. We might be left with an instrumental track after canceling. Then, we could use the 'inverse' of instrumental extract to isolate vocals from the audio track.

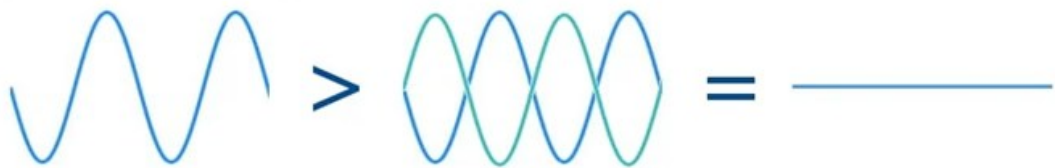

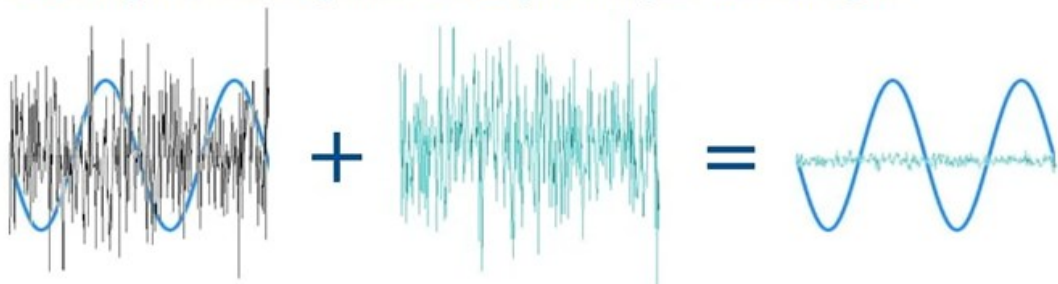

Figure 8: Idea of mixing audio with inverted noise, image by Luca Masella / Zaniboni (ref. 17).

Below is the Python code that was applied for canceling out the center using the PyDub library:

```
# [CANCEL CENTER: APPROACH 2.]
from pydub import AudioSegment
from pydub.playback import play

# Read in audio file and get the two mono tracks
sound_stereo_segment = AudioSegment.from_file("pantterinousut.wav",
format="wav")
sound_monoLeft = sound_stereo_segment.split_to_mono()[0]
sound_monoRight = sound_stereo_segment.split_to_mono()[1]

# Invert phase of the Right audio file
sound_monoR_inv = sound_monoRight.invert_phase()

# Merge two L and R_inv files, this cancels out the centers
sound_CentersOut = sound_monoLeft.overlay(sound_monoR_inv)
```

```python
# Write merged audio file where ideally vocals are removed
fh = sound_CentersOut.export("pantterinousut_centersout.wav",
format="wav")

# Invert sound_CentersOut
sound_CentersOut_inverse = sound_CentersOut.invert_phase()

# Phase cancellation on original stereo with inverted sound_CentersOut,
# this should isolate vocals.
vocals = sound_stereo_segment.overlay(sound_CentersOut_inverse)
fh = vocals.export("pantterinousut_vocals.wav", format="wav")
```

**Approach #3: State of the Art stem splitter using LALAL.AI**

We do not know the architecture of LALAL.AI (ref. 12), but it is likely similar to a baseline model Coffey and Pugh (ref. 16) have been working on as depicted in Figure 9. We just uploaded our audio track for processing by LALAL.AI and downloaded the separated vocals and instrumental tracks. Finally, we applied the speech-to-text algorithm to the vocals track.

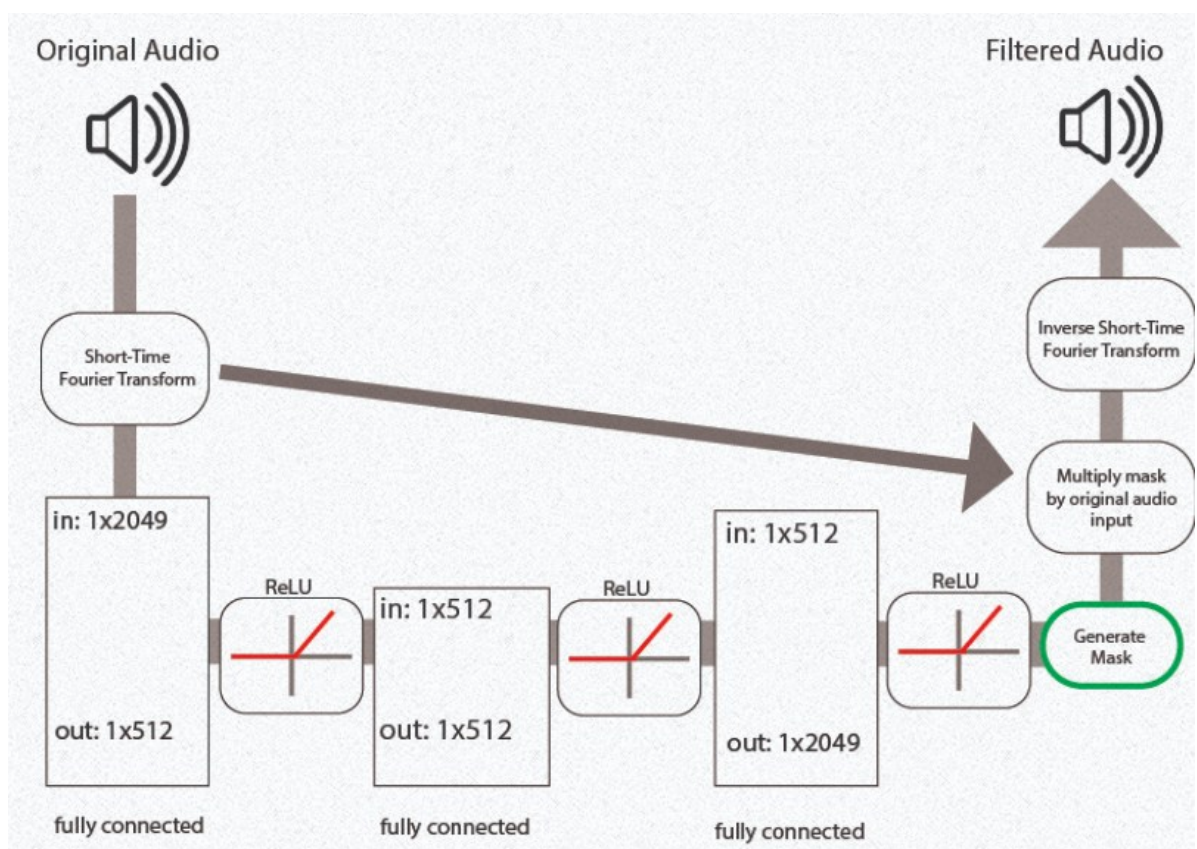

Figure 9. Machine learning model for stem splitter from Coffey and Pugh, ref. 16.

## 8. Re-evaluation of Hallucination Levels with pre-processed data

Next, we hand-pick some parts of the Finnish rap lyrics from the Pantterinousut song and compare the following two methods:
- Method A "Raw audio": Faster Whisperer model applied to original audio file (which has instrumental background music), and

- Method B "Pre-processed audio": Faster Whisperer is applied to the vocal track extracted from the original audio file using LALAL.AI (stem splitter algorithm).

Finally, we evaluate whether approach A or B wins per each selected lyric. Again, the paper's first author is going to be the judge here.

*Lyric #1*
Real: **"…Kuka tykkäs JRstä oli täys pelle"**
A/Raw: "Kuka tykkäs nii järjestä oli täys pelle."
B/Pre-processed: "Kuka tykkäs JRstä oli täys pelle."
Verdict: Method B wins (perfect extract), whereas A hallucinates.

*Lyric #2*
Real: **"MacGyveri ja Saloran töllöstä"**
A/Raw: "MacGyveri ja Saloran töllöstä"
B/Pre-processed: "Mäkkaai veriassaloran töllöstä"
Verdict: Method A wins (perfect), whereas B hallucinates.

*Lyric #3*
Real: **"Lakki lukei lähikirjastosta"**
A/Raw: "Lakki lukkee lähikirjastosta"
B/Pre-processed: "Latki lukee lähikirjastosta"
Verdict: Both mishear method A having Levenshtein distance 2 and B has distance 2 → It is a draw.

*Lyric #4*
Real: **"Skeletori bounssaa" (or "Skeletori baunssaa")**
A/Raw: "Skeletori pausaa"
B/Pre-processed: "Skeletori paunssaa"
Verdict: Both mishear a bit, A has Levenshstein distance 4 and B 2
→ B wins.

*Lyric #5*
Real: **"Tiikeritankkeihin kenguru bensaa"**
A/Raw: "Tiikeritankkeihin kenguru bensaa"
B/Pre-processed: "Tiikeritankkeihin kenguru pensaa"
Verdict: Method A wins, whereas B mishears one character (Levenshstein distance is 1 but it is halved to ½ because close phonemic distance: b->p).

*Lyric #6*
Real: **"Supermarioitaan Nintendolla"**
A/Raw: "Supermari joitaan Nintendolla"
B/Pre-processed: "Supermarinoitaan Nintendolla"
Verdict: Both mishear a bit, A has Levenshstein distance 2 and B 1
→ Method B wins.

*Lyric #7*
Real: **"Pantteri nousuista pantterilaskuihin piripintaan"**
A/Raw: "Panteri nousuista panterilaskuihin piripintaan"
B/Pre-processed: "Pantteri nousuista pantteri laskuihin piripintaan"
Verdict: Method B is almost perfect (Levenhstein distance 1), whereas A mishears twice (A has Levenshstein distance 2).

The overall verdict is that method A "use raw input audio" won twice out of seven times. In contrast, method B "use pre-processed data with vocals extracted from original raw input audio" won three out of seven times. However, there is no statistical confidence in whether method A or B is the winner. Pre-processing of data (removal of instrumental music) improves text extracts for some parts of the lyrics, whereas for other parts, it makes them worse.

## 9. Conclusion and Future Work

We have studied how AI models work in speech-to-text tasks with the Finnish rap song. The task is not trivial. It is not clear that YouTube's algorithm performs better. We also tried three preprocessing approaches to help AI. Two of them failed (independent component analysis and using the canceling center approach) in improving the input signal for the model. However, the last approach using LALAL.AI stem splitter isolated the vocal track from the audio track. Even it did not decrease AI hallucination level in general.

There are some ideas on how to improve hallucination analysis. First of all, for a more thorough comparison between models, one could apply the following ASR models: *i*) wav2vec2 Finnish NLP model in Hugging Face (ref. 19), *ii*) AaltoASR (ref. 20), or *iii*) Commercial services such as Google Cloud Speech to Text, Amazon Transcribe, and Microsoft Azure Speech to Text (which all support the Finnish language). Secondly, we studied just one rap song and sampled its lyrics in hallucination evaluation. So, that does not give any statistical evidence (sample size n=1). But we have a gut feeling about finding that the YouTube algorithm is better than Faster Whisperer. Thirdly, somebody could automate our error function. However, that likely would require AI (e.g., ChatGPT, for example). It would allow easy and fast evaluation of multiple rap songs and multiple parts of lyrics. Fourthly, somebody could try more preprocessing methods. Namely, we can think of at least the following two: *i*) Applying a low-pass filter to the audio (yet, it is likely not going to separate instrument music from vocal in our case – but it might still help ASR a bit), and *ii*) Transforming audio($t$) time-series to complex domain $a(t)+\mathbf{i}b(t)$ using Fourier transform, and then performing the independent component analysis (ICA) in the frequency domain, and finally inverse transforming extracted latent signals to the original domain. Perhaps the FourierICA method can be used directly, ref. 22 and 23.

Finally, this has likely been the funniest semi-scientific article we have ever written during our careers. In addition, thank you for getting this far and not churning earlier. Figure 10 below includes some characters in the Pantterinousut video greeting you. However, you cannot feel the energy by watching the below static picture. We recommend you to listen and watch it online (ref. 1) with sounds on!

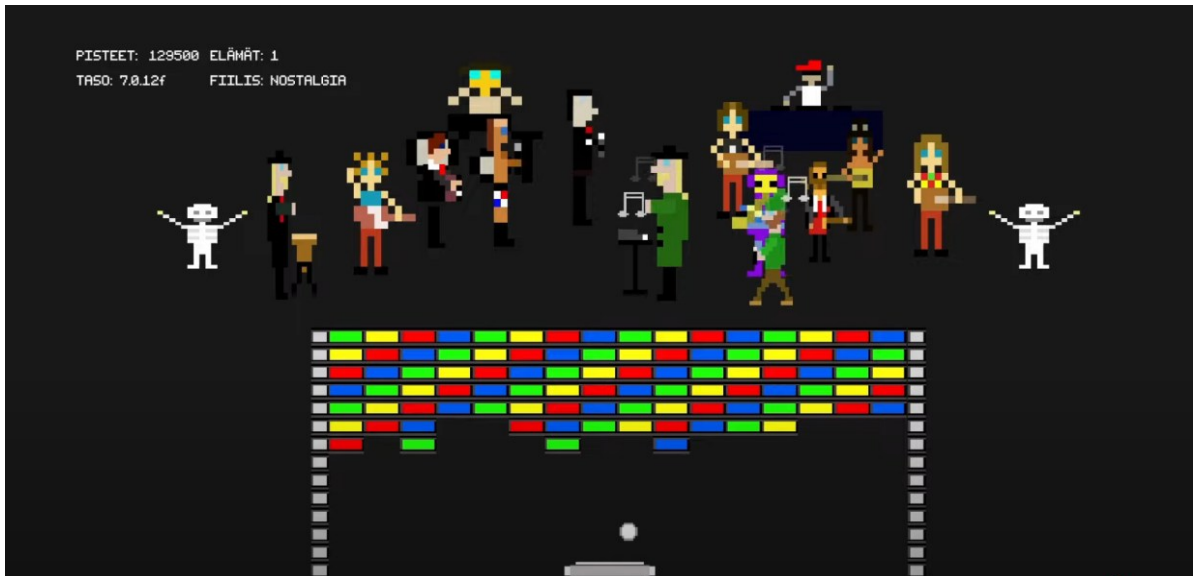
Figure 10. Characters from the Pantterinousut video.

**Acknowledgments**

We thank Omar Sanseviero for tweeting Figure 2 about the rules of thumb for selecting suitable Whisperer model complexity. Enes Zvorničanin for letting us use the independent component analysis (ICA) cocktail party chart (Figure 7). Luca Masella and Zaniboni for the picture about noise cancellation (Figure 8). Coffey and Pugh for their interesting work on stem splitters (Figure 9). Finally, this paper would not exist without the creative work done by Mc Timo and Syntikka Janne. Namely, we likely would not have been motivated to do anything like this with any other song.

**References**

1. Mc Timo & Syntikka Janne (2025). Pantterinousut. Internet url: https://www.youtube.com/watch?v=iq0hrThG96o (referenced on June 19[th] 2025).
2. Mc Timo & Syntikka Janne (2025). Facebook pages. Internet url: https://www.facebook.com/profile.php?id=61576624451346 (referenced on June 19[th] 2025).
3. Faster Whisper standalone Windows version, https://github.com/Purfview/whisper-standalone-win (referenced on June 19[th] 2025)
4. VLC Media Player with automatic AI subtitles, https://www.facebook.com/groups/726035349513216/posts/vlc-media-player-has-just-introduced-free-al-subtitles-with-real-time-translatio/999539558829459/ (referenced on June 19[th] 2025).
5. VLC Experimental Builds, https://nightlies.videolan.org/ (referenced on June 19[th] 2025)
6. OpenAI's Whisper, https://openai.com/index/whisper/ (referenced on June 19[th] 2025)
7. LLPlayer, https://github.com/umlx5h/LLPlayer (referenced on June 19[th] 2025)
8. PotPlayer, https://potplayer.daum.net/ (referenced on June 19[th] 2025)
9. Omar Sanseviero, "Which Whisper To Use" https://x.com/osanseviero/status/1725122881384776023 (referenced on June 19[th] 2025).
10. VLC Media Player, https://www.videolan.org/vlc/ (referenced on June 19[th] 2025)

11. Ans, B., Hérault, J., & Jutten, C. (1985). Architectures neuromimétiques adaptatives : Détection de primitives. Cognitiva 85 (volume 2, pp. 593-597). Paris: CESTA.
12. LALAL.AI, https://www.lalal.ai/ http, (referenced on June 19th 2025)
13. Taskinen, S., Sirkiä, S., & Oja, H. (2007). Independent component analysis based on symmetrised scatter matrices. Comput. Statist. Data Anal., 51(10), pp. 5103-5111
14. Ollila, E. (2010) Contributions to independent component analysis, sensor array and complex valued signal processing. Helsinki University of Technology. Doctoral Dissertation.
15. Ollila, E., Oja H., Koivunen V., (2008) "Complex-valued ICA based on a pair of generalized covariance matrices". Computational Statistics & Data Analysis, volume 52, number 7, pp. 3789-3805
16. Benji Pugh and Caitlin Coffey, https://caitlincoffey.com/finalprojectml/ (referenced on June 19th 2025).
17. Masella L. (2025) https://zaniboni.com/blog/post/noise-cancellation (referenced on June 19th 2025)
18. Audacity, https://www.audacityteam.org/ (referenced on June 19th 2025)
19. Hugging Face, https://huggingface.co/Finnish-NLP/wav2vec2-xlsr-300m-finnish-lm (referenced on June 15th 2025)
20. AaltoASR, https://github.com/aalto-speech/AaltoASR (referenced on June 19th 2025)
21. Fabian Berg, https://www.quora.com/Why-is-it-so-hard-to-hear-the-difference-between-p-b-and-t-d-Arent-people-just-saying-it-wrong-all-the-time (referenced on June 19th 2025)
22. A. Hyvärinen, P. Ramkumar, L. Parkkonen and R. Hari, (2010). Independent component analysis of short-time Fourier transforms for spontaneous EEG/MEG analysis, NeuroImage 49(1): pp. 257-271
23. FourierICA, https://www.cs.helsinki.fi/group/neuroinf/code/fourierica/html/fourierica.html (referenced on June 19th 2025).
24. Why Captions are Critical for YouTube, https://scribie.com/blog/2019/02/why-captions-critical-youtube/ (referenced on June 19th 2025).
25. Enes Zvorničanin https://www.baeldung.com/cs/independent-component-analysis (referenced on June 19th 2025).